\pdfoutput=1

\documentclass[11pt]{article}
\usepackage{authblk}
\usepackage[]{emnlp2021}

\usepackage{times}
\usepackage{latexsym}
\usepackage{todonotes}
\usepackage{latexsym}
\usepackage{booktabs}

\usepackage{multirow}
\usepackage{multicol}
\usepackage{graphicx}
\usepackage{subcaption}

\usepackage[T1]{fontenc}

\usepackage[utf8]{inputenc}

\usepackage{microtype}

\newcommand{\sharc}{{ShARC}}
\newcommand{\sharcstar}{{ShARC-PC}}
\newcommand{\zsdata}{{QA4PC}}
\newcommand{\maybe}{{Not Enough Information}}
\newcommand{\nei}{{NEI}}

\title{Cross-Policy Compliance Detection via Question Answering}

\author[1]{Marzieh Saeidi}
\author[1]{Majid Yazdani}
\author[1,2]{Andreas Vlachos}
\affil[1]{Facebook AI}
\affil[2]{University of Cambridge}
\affil[ ]{\texttt{\{marzieh,myazdani\}@fb.com, av308@cam.ac.uk}}

\begin{document}
\maketitle
\begin{abstract}
Policy compliance detection is the task of ensuring that a scenario conforms to a policy (e.g.\ a claim is valid according to government rules or a post in an online platform conforms to community guidelines). This task has been previously instantiated as a form of textual entailment, 
which results in poor accuracy due to the complexity of the policies. 
In this paper we propose to address policy compliance detection via decomposing it into question answering, where questions check whether the conditions stated in the policy apply to the scenario, and an expression tree combines the answers 
to obtain the label. Despite the initial upfront annotation cost, we demonstrate
that this approach results in better accuracy, especially in the 
cross-policy setup where the policies during testing are unseen in training. In addition, it allows us to use existing question answering models pre-trained on existing large datasets. 
Finally, it explicitly identifies the information missing from a scenario in case policy compliance cannot be determined. 
We conduct our experiments using a recent dataset consisting of government policies, which we augment with expert annotations
and find that the cost of annotating question answering decomposition is largely offset by improved inter-annotator agreement and speed.
\end{abstract}



\section{Introduction}

\begin{figure}[t]
    \centering
    \includegraphics[width=0.9\linewidth]{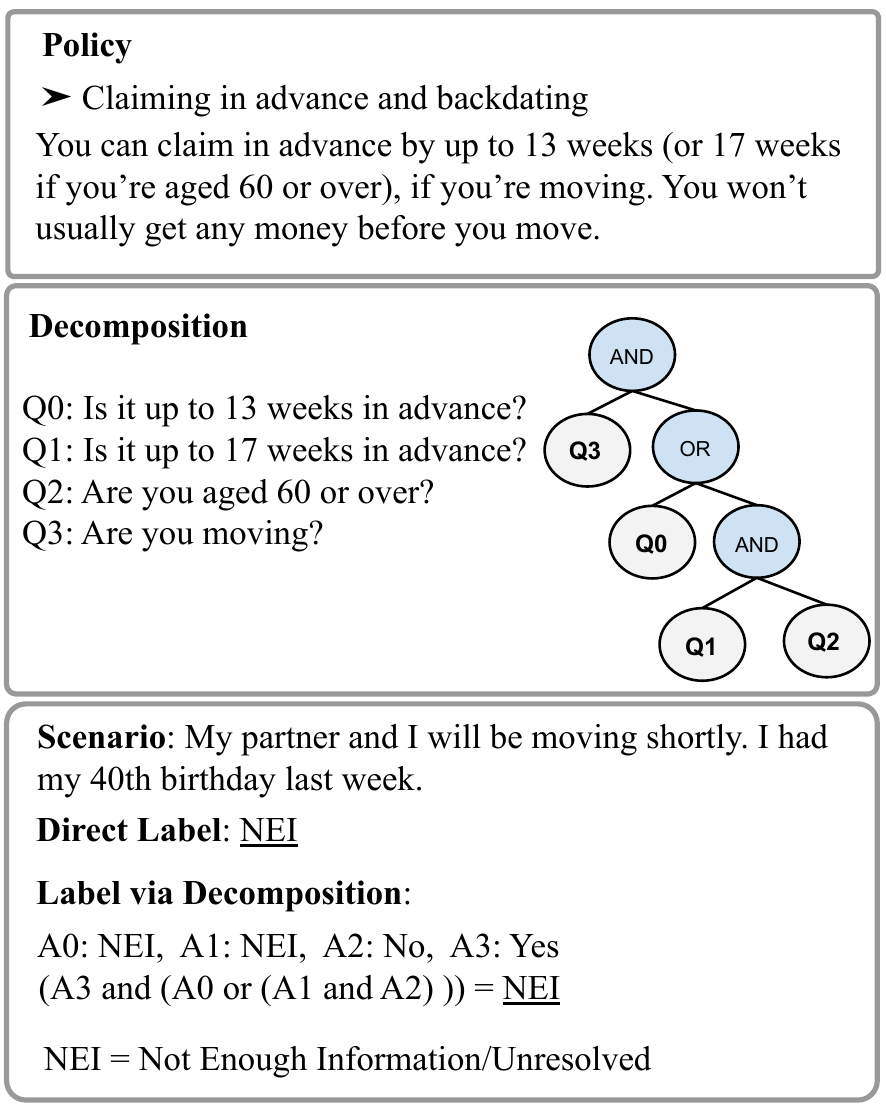}
    \caption{A policy with its decomposition (questions and expression tree), scenario and label.\protect\footnotemark}
    \label{fig_policy}
\end{figure}
 
\footnotetext{Policy source can be found here \url{https://www.gov.uk/housing-benefit/how-to-claim}}
Policy compliance detection is the task of ensuring that a scenario conforms to a policy. It is a task that occurs in various contexts, e.g.\ in ensuring correct application of government policies \citep{saeidi2018interpretation,jhu2020law}, enforcing community guidelines on social media platforms \citep{waseem-hovy-2016-hateful} or the resolution of legal cases \citep{zhong2019jec}, etc. 

While the task can be modelled as a text-pair classification similar to textual entailment~\citep{Dagan2009RecognizingTE,Bowman2015ALA}, i.e.\ whether a scenario complies with a certain policy, this often results in poor accuracy due to the complexity of the policy descriptions, which often contain multiple clauses connected with each other~\cite{jhu2020law}. 
Moreover, even though some types of policy compliance detection can be tackled via supervised classification, e.g.\ recognizing hateful memes \citep{kiela2020hateful} and other forms of hate speech or abusive language on social platforms \cite{hateoffensive}, this requires substantial amounts of training data for one specific policy which is expensive and time-consuming.
Furthermore, the requirement for policy-specific training data 
 renders supervised classification difficult to adapt when policies change, e.g.\ customs rules changing with Brexit, novel forms of hate speech in the context of the Covid pandemic, etc.

In this paper we propose to address policy compliance detection via decomposing it to question answering (QA) and an expression tree. Each policy description is converted into a set of questions whose answers are combined in a logical form represented by the expression tree to produce the label. This allows for better handling of complex policies, and to take advantage of the large-scale datasets and models developed for question answering \citep{rajpurkar2016squad, clark2019boolq}.
In addition, the questioning of the scenario with respect to parts of the policy identifies which parts were used in order to determine compliance and which information is missing in case the scenario could not be resolved. This in turn, enhances the interpretability of the model's decisions and can identify areas for improvement.
Figure~\ref{fig_policy} shows an example application of our approach in the context of a government policy where the missing date of the move in the scenario results in its compliance with the policy being unresolved.

We demonstrate the benefits of policy compliance detection via QA using a dataset that contains policies, decomposition questions and expression trees and scenarios. While the policies and scenarios in the dataset are taken from \sharc~\cite{saeidi2018interpretation}, we augment them with expression trees and answers to each question for all the scenarios and policies to create the Question Answering for Policy Compliance (\zsdata\ ) dataset.
The results of our experiments demonstrate that we can achieve an accuracy of $0.69$ for policies unseen during the training (an increase of $25$ absolute points over the entailment approach) and an accuracy of $0.59$ (an increase of $22$ absolute points) when no in-domain data is available for training. We also show that our approach is more robust compared to entailment when faced with policies of increasing complexity.
In addition, we find that the cost of annotating question answering decomposition and expression trees is largely offset by improved inter-annotator agreement and speed. 
Finally, we release the \zsdata\ dataset to the community to facilitate further research.\footnote{Please contact the authors to access the dataset.}

\section{Previous Work on Policy Compliance} \label{sec_policy_related_work}
Policy compliance as an entailment task has been studied by~\citet{jhu2020law} and referred to as statutory reasoning entailment. They find that the application of machine reading models 
exhibits low performance, whether or not they have been fine-tuned to the legal domain, due to the complexity of the policies. As an alternative approach, they propose a rule-based system. While their work focuses solely on the US Tax Law, the dataset we use in our experiments~\cite{saeidi2018interpretation} contains a wide range of policies from different countries.
Furthermore, the dataset used in~\cite{jhu2020law} only contains scenarios that are either compliant or not compliant with a policy. However, in real world use,
systems should be capable of recognizing that a decision cannot be made in cases there is information missing from the scenario.

\sharc~\cite{saeidi2018interpretation} is a dataset 
for conversational question answering based on government policies. In \sharc, the task is to predict what a system should output given a policy and a conversation history, which can be an answer or a follow-up clarification question. 
The task is evaluated in an end-to-end fashion, measuring the system's ability to provide the correct conversation utterance, 
However, policy compliance detection is neither investigated nor evaluated. 

Similarity search or matching has been used to match a new post to 
known violating posts
on social media such as hoaxes, an objectionable video or image, or a hateful meme~\cite{ferreira2016emergent, wang2018relevant,fbmatching}. 
For textual content, this can be compared to an entailment task. 
This approach requires a bank of existing violating content for each policy. 
By using the policy description and decomposing it into several QA tasks, breaches for new policies can be detected, as we show in our experiments.

Much work has focused on learning important elements of privacy policies to assist users in understanding the terms of the policies devised by different websites. \citet{shvartzshanider2018recipe} uses question answering to extract different elements of privacy policies that are informative to the users. \citet{nejad2020establishing} and \citet{mustapha2020privacy} assign pre-defined categories to privacy policy paragraphs using supervised classification. While these works aim to help users in understanding complex text of privacy policies, they do not aim to identify compliance and they mainly focus on privacy policies. In our work, we look at a wide range of government policies, for different governments and focus on an approach for 
cross-policy compliance detection.

\section{Policy Compliance Detection via QA}
In this section we describe our proposed approach for policy compliance detection via question answering. First we formulate the task as textual entailment following previous work~\citet{jhu2020law}. Then we describe how our proposed expression trees combined with QA operate. Finally we discuss the implications of our approach for training and data annotation requirements.

\subsection{Policy Compliance Detection (PCD)}

We define the task of policy compliance detection (PCD) as deciding whether a scenario is compliant with a policy.
For a policy and a scenario, the task is to provide a label in [\emph{Yes}, \emph{No}, \emph{\maybe}]. We use \nei\ in short for \maybe\ for the remaining of the paper. For an input $(p, s)$, the output is the label $l$. 
 See Figure~\ref{fig_policy} for an example.
If framed as textual entailment, a policy is the premise and the scenario is the hypothesis, as there are typically multiple scenarios per policy. The labels [\emph{Yes}, \emph{No}, \emph{\nei}] correspond to [\emph{entailment}, \emph{contradiction}, \emph{neutral}]. 

\subsection{Question Answering for PCD}
Our proposed decomposition of policy compliance detection into question answering infers the label using the answers to the questions and the expression tree. More formally, the QA model will take an input  $(s, q_i)_{i=1 \ldots K-1}$ and output $a_i \in$ [\emph{Yes}, \emph{No}, \emph{\nei}], where $K$ is the number of questions representing the policy $p$. The expression tree combines the answers to the questions, $\{a_i\}_{i=0 \ldots K-1}$, using a logical expression based on the rules in the policy description to infer the final label $l$ of whether a scenario is in compliance with the policy or not. Figure~\ref{fig_policy} shows an example of a policy decomposed into questions and an expression tree and how it is applied to a scenario to infer a label.

An expression tree can contain OR, AND and NOT operators. Expression trees can be evaluated in the same way that logical expressions are evaluated where \emph{Yes} is considered as True and  \emph{No} as False. Similar to logical expressions, we can evaluate an expression tree even when the answers to some of the questions are \nei. For example, the inferred label for ``Q0 AND Q1'' is False if the answer to Q0 is False and the answer to Q1 is \emph{\nei}.
If the answer to Q0 is True and the answer to Q1 is \emph{\nei}, the inferred label is \emph{\nei}.
It is worth noting that we do not always need to correctly predict the answers to all the questions in the expression tree in order to infer the correct label. For example, for the expression tree in Figure~\ref{fig_policy}, if the answer to Q3 is correctly predicted as No, the model doesn't need to make correct predictions for Q0, Q1 or Q2.

The underlying QA task is similar to BoolQ~\cite{clark2019boolq} with the exception that an answer can be \nei. This can be compared to SQUAD~2.0~\cite{rajpurkar2018know} where the corpus contains unanswerable questions, but the answers are spans extracted from the passage instead of boolean. In this work, we assume that the expression trees are provided and leave the task of  inferring them for future work. 

\subsection{Training and Data Requirements}
Expression trees are not required during the training of our approach; they are only used during the testing to infer the label for the scenario given the policy. 
For this reason, we only provide the annotation of expression trees for the data that will be used for evaluation (dev and test sets) to demonstrate the benefits of our proposed approach. Using QA as the subtask, in which we decompose PCD into, enables us to apply our approach to policies unseen in training (a.k.a.\ cross policy 
setup), since QA models are typically developed for and evaluated on their generalization ability to questions unseen in their training data.

\section{Data}\label{sec:data}
In this section we discuss the creation of the question-answering for policy detection (\zsdata) dataset which we introduce in this work to support our experiments.
We create \zsdata\ using the policies and the scenarios from the publicly available training and development splits of existing dataset \sharc~\cite{saeidi2018interpretation}. A scenario in \sharc\  is a real-world situation described by a user who is conversing with a system to find out whether they comply with a policy.
Each conversation utterance in \sharc\ has a policy, a question and an answer. An utterance may include a  conversation history (a list of QAs) and/or a scenario which is built from a conversation list (a list of questions and answers). The answer to an utterance can be Yes/No or a follow-up clarification question.~\footnote{More details on \sharc\ data can be found here \url{https://sharc-data.github.io/data.html}} 

However, while \sharc\ conversations contain some of the questions related to the policies, we found that many were missing, as the conversation progress may render answers to them irrelevant. A such example is provided in Figure~\ref{fig_sharc_less_qa} in the Appendix. Details on number of QAs per policy for both datasets are discussed in section~\ref{sec_zsdata}.

 As neither entailment-style classification nor the question answering expression trees require complete question-answer sets per instance or expression trees for training, we convert $70\%$ of \sharc\ policies for training of entailment and QA tasks (see section~\ref{sec_in_domain_ds} for more details). The procedure for converting the data is explained in the next section. In order to ensure the evaluation datasets (dev and test sets) are correctly annotated, we augment the remaining $30\%$ of \sharc\ policies with questions and expression trees, and the scenarios with the entailment labels and answers to all questions of the related policy (see Section~\ref{sec_annotation}). 
Two annotators, co-authors of this paper and UK nationals, were involved in the annotation described in this section.

\subsection{Converting \sharc\ to PCD}

\paragraph{Entailment Instances} We take utterances that have non-empty scenarios. The entailment answer to the scenario is Yes/No if the answer to the conversation utterance in \sharc\ is Yes/No and the conversation history is empty. Conversation history refers to a list of questions and answers that has already been exchanged between an agent and a user. Otherwise, the assigned label is \nei. This is because if the conversation history is not empty or the final answer is a follow up question, some of the necessary information related to the policy is not mentioned in the scenario and is acquired from history or will be answered by the user in the next step of the conversation in the form of an answer to the follow up question.

\paragraph{QA Instances}
We construct the set of unique follow-up questions related to each policy over all the scenarios as the set of questions for that policy.
If any question from this set appears in the conversation corresponding to a scenario (a list of QAs), we create a QA instance where the passage is the scenario, and the answer is the answer to that question in the QA list. For all the other questions in this set, we add a QA instance using the scenario and the answer \nei. More details about the \sharc\ dataset and its processing are included in the Appendix in section~\ref{sec_app_sharc}.

\paragraph{In-Domain Supervision} \label{sec_in_domain_ds}

We call the converted \sharc\ dataset \sharcstar. There are $482$ policies, $4,576$ scenarios and $10,398$ QAs in the \sharc\ training set. The policies used in \sharcstar\ are distinct from those used in \zsdata. Note that the labels in \sharcstar\ are not always accurate because they are assigned using a heuristic, thus even though it is in-domain as it is derived from conversation about government policies, it is noisy. This was specially noted when augmenting the \sharc\ data to obtain \zsdata.

\subsection{Expression Tree Annotation}
\label{sec_annotation}
\paragraph{Annotation Procedure} 
Each policy is decomposed by the annotators into a set of questions and an expression tree based on the rules in the policy description. Each question is assigned an ID (e.g. Q0, Q1 ) which is used in the expression tree. For each policy, some of the questions can be extracted from the \sharc\ dataset as mentioned in the previous section. These questions are provided to the annotators. An annotator can change the question or remove it. Annotators can also add questions to expression tree if the existing questions are not sufficient. The UI for this annotation task is depicted in Figure~\ref{fig_annotation_tree_ui} in the appendix. It took the annotators an average of $113$ seconds to annotate each tree.

\begin{figure*}[!t]
\centering
\begin{subfigure}{.45\textwidth}
  \centering
  \includegraphics[width=.65\linewidth]{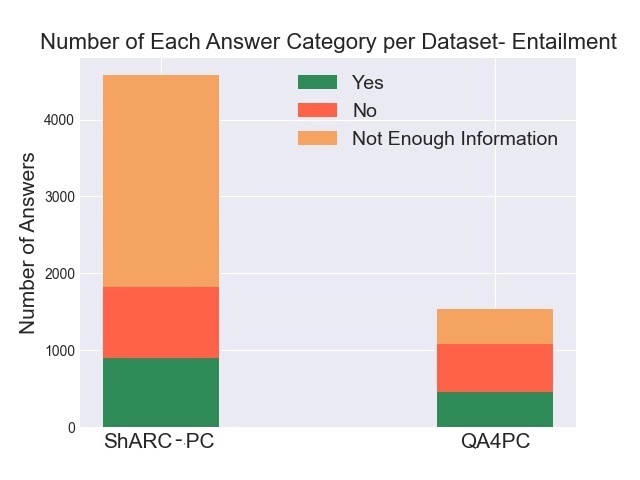}
\end{subfigure}%
\begin{subfigure}{.45\textwidth}
  \centering
  \includegraphics[width=.65\linewidth]{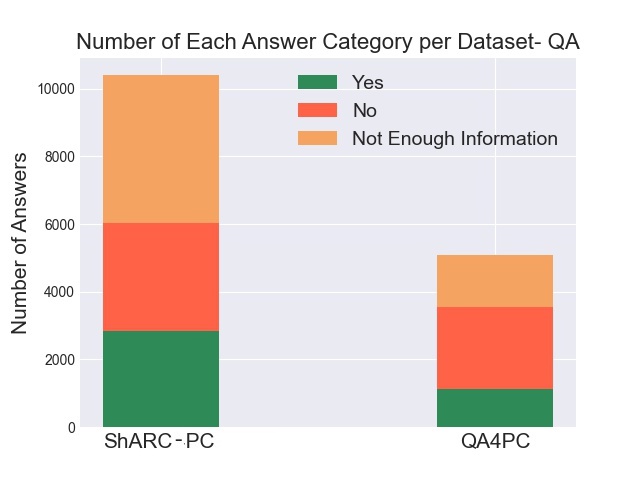}

\end{subfigure}
\caption{Label/answer types for training and evaluation datasets for entailment (left) and QA (right) tasks.}
\label{fig_answer_types}
\end{figure*}

\paragraph{Quality Control}
Annotators were asked to mark difficult instances and also take note during the annotation if they were in doubt about an instance. After the first round of annotation, the annotators discussed their notes and clarified the guidelines. 
Finally, each annotator reviewed the other annotator's expression trees and made updates if necessary based on the interim discussion. To further ensure the quality of the expression trees, we asked a third annotator (third co-author, not involved in the annotation otherwise) to review a sample of $50$ trees. These trees were selected proportional to their complexity, i.e.\ number of questions and logical operators. Out of $50$ trees, the third annotator did not agree with $4$ annotations, i.e.\ an agreement of $92\%$. The annotator marked $6$ additional instances where they believed the framing of questions can be improved.

\subsection{Entailment and QA Annotations}
\paragraph{Annotation Procedure}
Each scenario is paired with its corresponding policy. 
An annotator is required to choose a label from (Yes/No/\nei) with regards to the policy compliance (i.e. entailment instance). In the same annotation task, the annotators are asked to provide answers to each individual question in the expression tree of the policy (QA instances). The UI for this task is shown in Figure~\ref{fig_annotation_sc_qa_ui} in the appendix.

\paragraph{Quality Control}
We compared the inferred labels to all scenarios using the expression tree of the corresponding policy against the entailment label provided as a sanity check. 
Annotators were also asked to take notes during the annotation if needed and discussed their notes and the discrepancies identified by the sanity check. In some cases, answers were adjusted appropriately. In other cases, the expression tree was updated, and the annotators ensured that the question and answers were updated accordingly.

\subsection{\zsdata\ Dataset} \label{sec_zsdata}
In this section, we present statistics of the \zsdata.
As neither our approach nor the entailment baseline require training data annotated with expression trees, we divide the dataset only into dev and test sets. There are $193$ policies in \zsdata\ across both test and dev sets. Each policy has between $1$ and $9$ questions and between $2$ and $55$ scenario instances.
Table~\ref{dataset_num_trees_scenarios_qas} shows more properties of dev and test sets such as the total number of scenarios and individual QAs and average number of QAs per policy. 
The published dataset contains the training, dev and test sets.

\begin{table}
\centering
\small
\begin{tabular}{l | llll}
 & \textbf{Policies} & \textbf{Scenarios} & \textbf{QAs} & \textbf{Avg QA/Policy}\\
\hline
\textbf{Dev} & $60$  & $437$ & $1,600$ & $2.31$\\
\textbf{Test} & $133$ & $1,099$ & $3,492$ & $2.30$\\
\end{tabular}
\caption{
Number of instances in \zsdata\ dataset.} \label{dataset_num_trees_scenarios_qas}
\end{table}

Policies in \zsdata\ have an average number of $2.31$ QAs. This is higher than the average number of QAs per policy in \sharcstar\ which is $1.73$, because annotators added additional questions to the policies during the annotation when required. Figure~\ref{fig_answer_types} shows the distribution of entailment labels and answers to the QA instances between training (i.e. \sharcstar) and evaluation (i.e. \zsdata) sets. As the figure shows, the training data has a higher number of instances where the entailment labels (left) or answers to the questions are \nei\ (right).

\begin{table*}[!t]
\small
\centering
\begin{tabular}{l | lll | lll}
& \multicolumn{3}{c}{\textbf{Avg over Scenarios}} & \multicolumn{3}{c}{\textbf{Avg over Policies}}\\
\hline
\textbf{Training data} & \textbf{\sharcstar}  & \textbf{BoolQ} & \textbf{SNLI} & \textbf{\sharcstar} & \textbf{BoolQ} & \textbf{SNLI}\\
\hline
& \multicolumn{6}{c}{\textbf{Test Set}} \\
\hline
\textbf{Entailment} & $0.44 \pm 0.02$ &  NA & $0.37 \pm 0.02$  & $0.44 \pm 0.03$ &  NA & $0.42 \pm 0.02$\\
\textbf{QA ET} & $0.69 \pm 0.01$  &  $0.59 \pm 0.02$ & NA & $0.69 \pm 0.02$  &  $0.62 \pm 0.02$ & NA\\
\hline
& \multicolumn{6}{c}{\textbf{Dev Set}} \\
\hline
\textbf{Entailment} & $0.44 \pm 0.01$ &  NA & $0.35 \pm 0.02$  & $0.48 \pm 0.01$ &  NA & $0.39 \pm 0.02$ \\
\textbf{QA ET} & $0.70 \pm 0.01$  &  $0.65 \pm 0.02$ & NA & $0.74 \pm 0.02$  &  $0.65$ & NA\\
\end{tabular}
\caption{Cross-policy accuracy 
averaged on all scenarios and per policy on test and dev sets in terms of macro-accuracy, averaged over 5 runs with varying seeds. QA ET refers to QA decomposition with Expression Trees.}
\label{tab_all_results}
\end{table*}

\section{Experimental Setup}
\paragraph{Model}
We use RoBERTa~\cite{liu2019roberta} for both the entailment baseline and the QA subtask and learn three-way classifiers for each. 
We used the implementation in the huggingface~\cite{wolf-etal-2020-transformers} library. We also experimented with a T5 model~\cite{raffel2019exploring} on the dev set. Since the results of T5 model were very similar to RoBERTa, we report the performance on all the tasks using RoBERTa.

\paragraph{Entailment}
We use a cross-encoder set up for the task. For each instance, we combine the policy, scenario and the question using the following format: ``premise: [policy] SEP hypothesis: [scenario] CLS''.
The embeddings of the token CLS is used to classify the instance. This is a common approach for entailment modelling using pre-trained encoders~\cite{devlin2018bert,raffel2019exploring, jhu2020law}.

\paragraph{Question Answering for PCD}
In the QA approach to PCD, we decompose the task into a QA subtask and expression trees, where we use the expression tree of a policy to combine the answers to the questions in the policy to infer the label.
A QA model is trained on QA instances of the dataset. 
For each QA instance, we combine the scenario and the question using the format: ``passage: [scenario] SEP question: [question] CLS''. This is a common formulation of QA tasks~\cite{raffel2019exploring} using transformer models.
The embedding of the CLS token  is used to assign a label to each instance.

\paragraph{Out-of-Domain Training Data}
We use the BoolQ~\cite{clark2019boolq} dataset for training the QA model. BoolQ only contains instances (passage and question pair) with Yes/No answers. We pair random passages and questions to create instances with \nei\ answers. To train the entailment model, we use the SNLI dataset~\cite{Bowman2015ALA}. To adapt the labels to our task, we convert entailment to Yes, contradiction to No and neutral to \nei.

\paragraph{Evaluation Metric}
We use macro-accuracy (averaged over 3 classes of PCD labels/QA answers) 
 averaged over scenarios, since the labels and answers are not balanced in our dataset (see Figure~\ref{fig_answer_types}). As some policies have more scenarios than others, we also report results  averaged over policies, where we first calculate the macro-accuracy per policy and then average over policies.

\paragraph{Hyperparameters}
 We did a manual tuning of hyperparameters using the dev set. Batch size was set to $16$, learning rate to $5e-5$, Adam epsilon to $1e-8$ and maximum gradient norm of $1.0$. We trained all the models for a maximum of 5 epochs with early stopping using the loss on the dev set. We run each experiment $5$ times with different random seeds and report the mean and variance. The models were ran on a machine with 8 Tesla V100-SXM2 GPUs, each with $16$ GB memory. Each epoch of model training takes about $25$ seconds for the entailment baseline and $35$ seconds for the QA subtask on \sharcstar. Number of trainable parameters of the model is $124,647,939$.


\section{Results}

\subsection{Model Accuracy} \label{sec_model_acc}
The accuracies of different approaches in detecting policy compliance and using different training data are presented in Table~\ref{tab_all_results}.
\begin{table}
\small
\centering
\begin{tabular}{l|ll}
 & \textbf{\sharcstar}  & \textbf{BoolQ} \\
\hline
\textbf{QA} & $0.68 \pm 0.02$ & $0.59 \pm 0.01$ \\
\end{tabular}
\caption{The performance of the QA task where models are trained on in-domain and out-of-domain data and evaluated on the test set of \zsdata.}
\label{tab_qa_results}
\end{table}

\begin{table}
\small
\centering
\begin{tabular}{l|lll}
 & \textbf{Yes}  & \textbf{No} & \textbf{\nei}\\
\hline
\textbf{QA} & $0.84 \pm 0.03$ & $0.54 \pm 0.05$  & $0.66 \pm 0.03$\\
\textbf{PCD} & $0.80 \pm 0.03$ & $0.68 \pm 0.04$  & $0.59 \pm 0.02$\\
\end{tabular}
\caption{The performance of QA and PCD tasks per label using the model trained on \sharcstar.}
\label{tab_labels_results}
\end{table}

The first column of the table shows the performance of both approaches in detecting policy compliance on unseen policies when trained on data from \sharcstar. The accuracy of the entailment approach is $0.44$ while the QA decomposition approach reaches an accuracy of $0.69$. Table~\ref{tab_qa_results} shows the performance of the QA subtask (macro-accuracy over QA instances) on \zsdata\ test set using in-domain (i.e. \sharcstar) and out-of-domain (i.e. BoolQ) training data. As mentioned previously, we do not need to answer all the questions in a policy correctly in order to get the correct final label for PCD. For example, in Figure~\ref{fig_policy}, if the answer to Q3 is correctly predicted as No, inferring a correct final label is independent of the remaining answers.
This is the reason that a QA model with an accuracy of $0.68$ can still achieve an accuracy of $0.69$ on the PCD task.

Since we have a class imbalance in our dataset, we show the accuracy of the model per label for both the QA and the PCD tasks in Table~\ref{tab_labels_results}. The most difficult answer for the QA model to predict is \emph{No}, while the answer with the lowest prediction accuracy for the PCD task is \emph{\nei}. 

\paragraph{Transfer Learning}
We define transfer learning as learning from a training set that is out-of-domain (i.e.\ not government policies). Table~\ref{tab_all_results} shows transfer learning results for the PCD task based on training on BoolQ for QA-based approach and SNLI for the entailment baseline. As we can see, the decomposition approach achieves an accuracy of $0.59$ when trained on BoolQ data. On the other hand, the accuracy of the entailment model trained on SNLI is $0.37$ which is only slightly better than a random system. This is likely due to entailment detection between policy and scenario being a more difficult task due to the complexity of the policy description, and it is not possible to obtain good accuracy without in-domain data.

\paragraph{Performance per Policy}

We evaluate the accuracy of both approaches averaged over policies to show how well we can infer compliance with a new policy. 
The results averaged over policies are shown in the right side of Table~\ref{tab_all_results}. The accuracies averaged over policies are similar or slightly higher than those averages over all the scenarios. This indicates that there are policies that are difficult to do inference on which have a high number of instances.
This is evident for both the entailment and QA decomposition approaches. 

\begin{figure}
    \centering
    \includegraphics[width=0.8\linewidth]{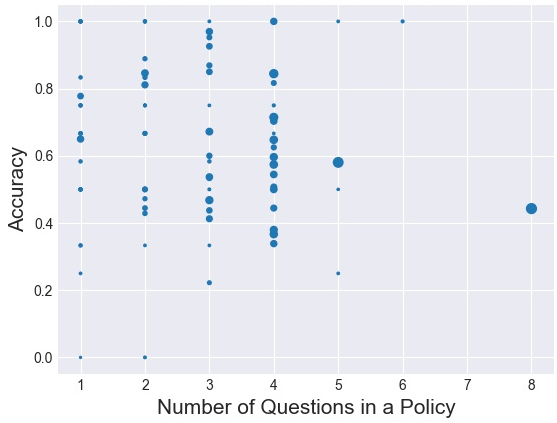}
    \caption{Performance (test set) per policy vs. the complexity of the policy, i.e.\ the number of questions in the expression tree.}
    \label{fig_acc_per_policy_size}
\end{figure}

Figure~\ref{fig_acc_per_policy_size} shows the average performance of the PDC task using QA decomposition approach for individual policies in the test set of \zsdata\ versus the complexity of the policy description, i.e.\ number of questions in the policy expression tree. The size of the circles indicate the number of examples in the dataset for a policy.
We used Kendall tau to find the correlation between the number of questions in a policy and the accuracy. Tau coefficient between accuracy and number of questions in policies using QA and entailment approaches are $-0.12$ (p-value of $0.05$) and $-0.17$ (p-value of $0.01$) respectively. This shows that the accuracy through the entailment approach is more negatively affected as the complexity of a policy increases.
 Finally, it is worth noting that a policy with $8$ questions has a high number of instances. The average accuracy over these instances is $0.42$ which contributes to the slightly lower performance when averaged over the scenarios reported in Table~\ref{tab_all_results}.

\paragraph{Limited Supervision}

\begin{figure}
    \centering
    \includegraphics[width=0.8\linewidth]{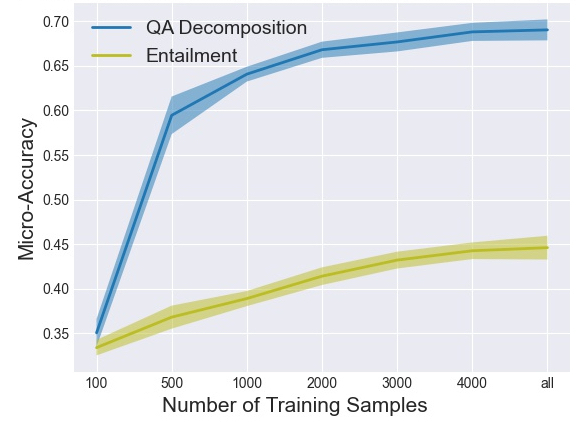}
    \caption{Performance of two approaches on the test set as the number of training samples (scenario) from in-domain data increases.}
    \label{fig_results_few_shot}
\end{figure}

In Figure~\ref{fig_results_few_shot}, we show the performance of the model on \zsdata\ as the amount of in-domain training data from \sharcstar\ increases. The results are shown in blue and green for the QA decomposition and the entailment approach respectively. As figure shows, with only $500$ scenario instances, we achieve an accuracy of $0.60$ through the QA decomposition approach. The entailment approach reaches an accuracy of $0.40$ using more than $1000$ instances. Note that the number of training instances are based on the number of scenario instances which can include a higher number of QA instances. However we will demonstrate the annotation time is not higher for more QA instances in the next section. Also, it takes only an average of $113$ seconds to annotate the expression tree (see section~\ref{sec_annotation}) which is done only once per policy and used for multiple scenarios.

\subsection{Annotation Accuracy and Efficiency} \label{sec_annotation_efficiency}
To perform policy compliance detection via QA decomposition, not only we need annotations for expression trees, we also need to annotate the answers to all the questions of a policy for a related scenario instance. Therefore, for each scenario, instead of needing one entailment label, we need annotators to provide answers to all the questions in the expression tree of a policy. One may argue that this approach is more costly as a higher number of annotations is required and that this effort could be used instead to annotate more data on the entailment level that can lead to equally good or better performance. In this section, we describe our experiment to demonstrate the effectiveness of the annotation for the QA decomposition approach in comparison to the entailment. 
\paragraph{Task Setup} 50 policies were selected from \zsdata. We include policies with different number of questions (and therefore complexity) in this experiment. For each policy, one scenario is selected randomly from the dataset. Policies are unique to ensure that the accuracy and speed of annotators are not affected by their familiarity with the policies. We then create two annotation tasks for each scenario-policy instance. In the first task, the annotators are presented with the policy and the scenario and they are required to choose one entailment label. In the second task, the scenario and \emph{all} the questions related to the policy are presented to the annotators. An annotator is required to provide an answer to each of the questions.

\paragraph{Agreement and Speed} We used 
two annotators (unpaid volunteers who are not authors of the paper but are native English speakers) to perform both annotation tasks.
The average time to provide an entailment label for a policy and a scenario is $39$ seconds across two annotators. The average time to answer \emph{all} the questions of a policy with respect to a scenario is $27$ seconds. The annotators highlighted the difficulty of annotating the entailment task because they had to create a mental breakdown of a policy while they found the QA task easier to perform as the policy is already decomposed into individual questions. The Cohen's Kappa for entailment agreement between the annotators is $0.38$ and for individual questions $0.49$. The agreement on the inferred labels using the expression trees based on the answers to the questions is $0.70$. This agreement is higher than the agreement for the QA annotation, because as we discussed in section~\ref{sec_model_acc}, not all the the questions need to be answered correctly to infer the correct final label. The results of this annotation experiment shows that annotating the QA labels has lower cost (faster) compared to annotating the overall entailment label, even though an annotator needs to complete more annotations.

\section{Discussion}

\paragraph{Assumptions} In this work, we assume that we can decompose  policy descriptions into independent questions that can be executed in parallel. 
However, designing independent questions is not always straightforward. 
E.g.\ it could be useful to have a question in an expression tree that is a follow-up on another question and therefore a QA model will have to be executed sequentially taking into account the earlier question in formulating the follow up one.

\paragraph{Potential for Automatic Generation of Expression Trees}
In our work, we assume that the expression trees are provided.
Providing an expression tree for a policy by those who create it is unlikely to be a substantial overhead, and it can lead to better policy definitions. In addition, in the case of policies on sensitive issues such as hate speech, it might be undesirable to have the expression tree and questions inferred automatically. Moreover, an expression tree for a policy is created only once but used multiple times. As discussed in the results section, the cost of creating expression trees when using the QA approach is offset by the 
gains in performance compare to the entailment approach. 

Nevertheless, on-boarding all the policies of an existing organisation can be time-consuming and can benefit from automation, and with human-in-the-loop approaches   we can ensure the accuracy of the trees. Finally, a potential benefit of automating the generation of expression trees is that it can be done jointly with the task of policy compliance detection, such that we generate trees that result in a higher accuracy on the downstream task.

\paragraph{Use of Task Descriptions}
The use of task descriptions was recently studied in~\citet{weller2020learning} to answer questions based on a passage. Our work is similar to this work since we use policy descriptions to learn PCD and conduct evaluation in a cross-policy 
setting. While they answer questions from passage descriptions, we focus on learning policy compliance given a policy and a scenario. Description of relations has been used in \cite{obamuyide2018zero} to perform zero-shot relation classification.

\paragraph{Question Decomposition}
Answering complex questions has been a long-standing challenge in natural language processing. Many researchers explored decomposing questions into simpler questions~\cite{wolfson2020break, min2019multi,ferrucci2010building,perez2020unsupervised}. In question decomposition, 
the objective is to convert a complex question into a list of inter-related sub-questions. While in question decomposition questions are generated automatically, in our work, we consider them given. On the other hand, the answers to the decomposed questions are combined in a more complex manner using an expression tree with logical operators, while in these works there is no attempt to combine the answers.   

\section{Conclusion}
In this work, we propose to address the task of policy compliance detection via decomposing a policy description into an expression tree. The expression tree consists of a set of questions and a logical expression that combines the questions to infer a final label. Our experiments show that compared to the existing entailment approach~\cite{jhu2020law}, QA decomposition results in a better model accuracy in a cross-policy 
setting using in-domain ($0.69$ vs. $0.44$) and out-of-domain ($0.59$ vs. $0.37$) training data. Furthermore, we show that while there is an upfront cost of annotation for expression trees, the cost of QA decomposition annotations is lower than the cost of annotating the entailment task while reaching a higher agreement. 

Future work can investigate ways to generate expression trees for policy descriptions automatically. Further, it will be beneficial to demonstrate that this approach is suitable in detecting policy compliance in other domains such as community standards implemented by social media platforms which requires annotation of relevant datasets.

\section{Broader Impact Statement}
Tools based on our method could potentially improve automated policy enforcement efforts and could also help individuals understand how certain policies may apply to their circumstances. The proposed method offer better interpretability and performance when little data in available. This can provide some insights on where a model may have made a mistake. 

We use an existing transformer-based model to show the effectiveness of our method. Transformer-based models are known not to be computationally effective. 
However, we use a pre-trained model and fine-tune it on a relatively small dataset. 
\bibliographystyle{acl_natbib}
\bibliography{references}
\clearpage
\appendix

\section{Appendices}
\label{sec:appendix}

\subsection{\sharc\ Data Processing}\label{sec_app_sharc}
\sharc\ dataset contains government policies from a number of countries including the UK and the USA. for each policy, several scenarios are provided. A scenario is a life situation described by a user who is interested to know whether they comply with the given policy. Scenarios were created by providing a full or partial list of questions and their answers from a path in the tree to the labelers. Labelers were then asked to write a life situation compatible with those questions and answers. Annotators were also told to feel free to provide extra context and information, even if not directly relevant to the provided questions. Note that many tree instances in \sharc\ do not have any related scenarios and only rely on interactive information seeking with a human to arrive at an answer. 

Even though a binary tree for each policy can be inferred from the \sharc\ instances, we encountered two issues with the data: 1) The trees were not always complete and some fundamental questions were missing. 2) Trees with only possible Yes/No for each node as done in \sharc\ are not expressive enough (or it's too complicated to express a policy fully in a binary tree) to capture the semantics of the policies in many cases. For example, binary trees do not capture OR operators very well where the answer to some questions are not required in order to infer the final answer. Without an accurate binary tree, we can not guarantee to infer a correct label given the answers to the granular tasks.

\begin{figure}
    \centering
    \includegraphics[width=0.6\linewidth]{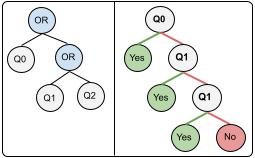}
    \caption{The logical tree of the expression ``Q0 OR Q1 Or Q2'' proposed in our work on the left and the decision tree proposed by ShARC.}
    \label{fig_sharc_vs_zs_trees}
\end{figure}

Figure~\ref{fig_sharc_vs_zs_trees} shows that binary trees in ShARC can not deal with the answers \emph{\maybe} whereas our proposed expression trees do.

\begin{figure}
    \centering
    \includegraphics[width=0.98\linewidth]{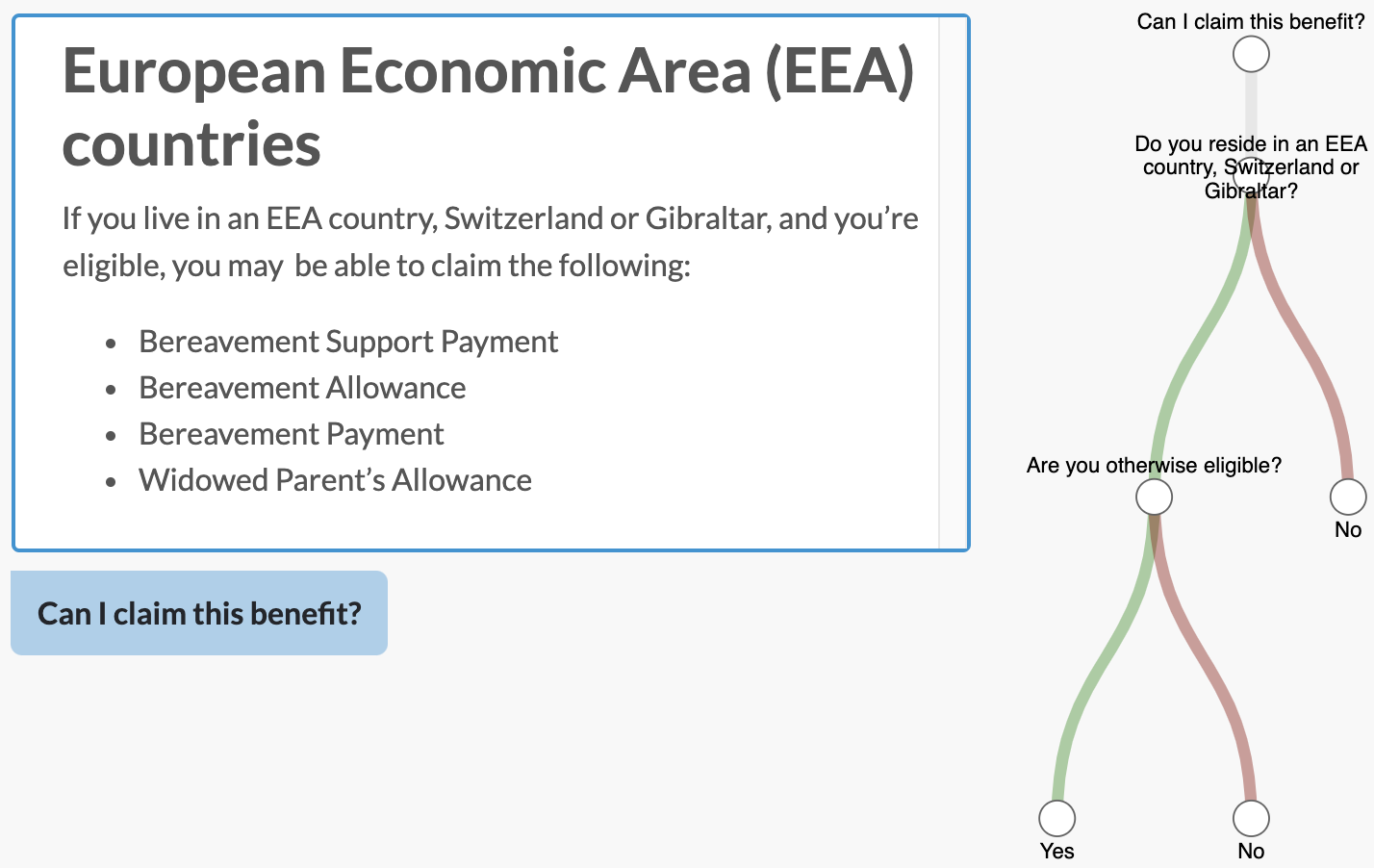}
    \caption{An example of a tree in ShARC where we added extra questions in \zsdata\ as we believed they were essential in evaluating a scenario. in \zsdata, four questions were added about the type of benefit corresponding to each bullet point.}
    \label{fig_sharc_less_qa}
\end{figure}

\subsection{Annotation}

\begin{figure}[h]
    \centering
    \includegraphics[width=0.95\linewidth]{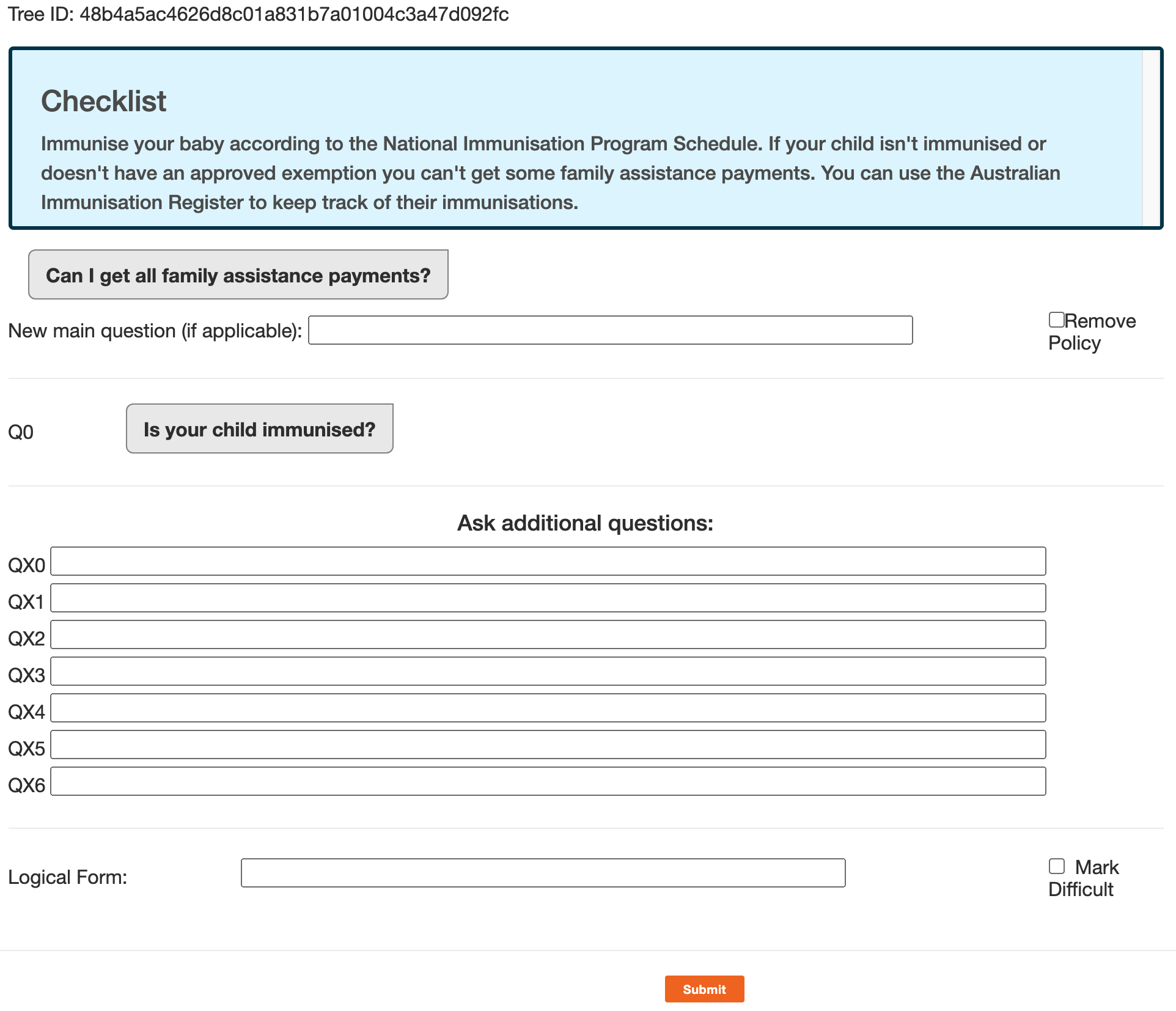}
    \caption{The annotation UI for expression tree annotations.}
    \label{fig_annotation_tree_ui}
\end{figure}
\begin{figure}[h]
    \centering
    \includegraphics[width=0.95\linewidth]{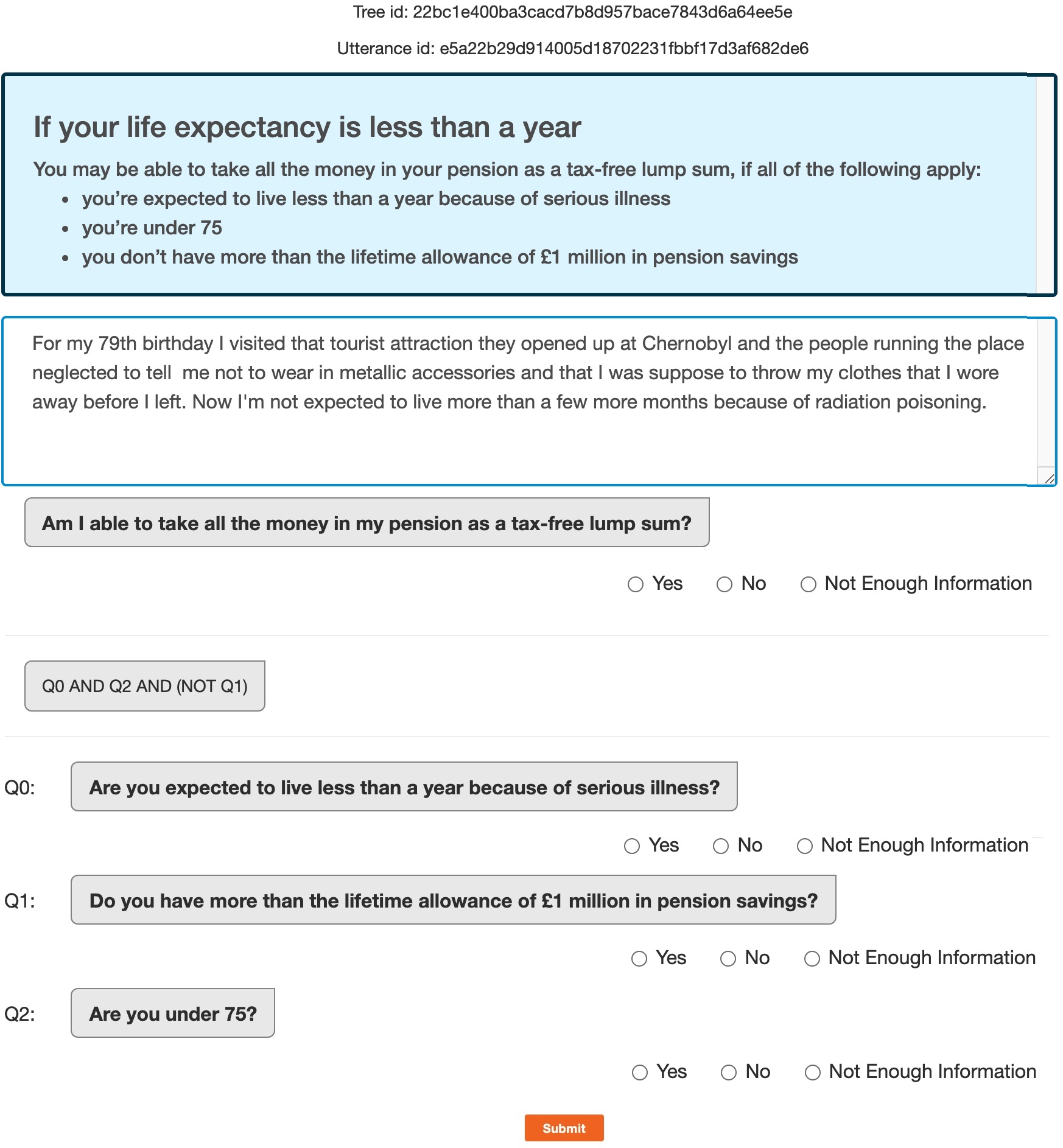}
    \caption{The annotation UI for QA and entailment tasks.}
    \label{fig_annotation_sc_qa_ui}
\end{figure}

\label{sec:supplemental}

\end{document}